\documentclass[runningheads]{llncs}

\usepackage{accv}

\usepackage{amsmath,amssymb} 
\usepackage{color}
\usepackage{cite}

\usepackage{booktabs}
\usepackage{colortbl}
\usepackage{pifont}
\usepackage{adjustbox}
\usepackage{threeparttable} 

\usepackage{siunitx}
\usepackage{tabularx} 
\usepackage{multirow} 
\usepackage{makecell} 

\newcommand{\cmark}{\textcolor{green!60!black}{\ding{51}}}%
\newcommand{\xmark}{\textcolor{red}{\ding{55}}}%

\usepackage{pgfplots}

\usepackage{mathtools}
\usepackage{tabularx} 
\usepackage{todonotes}

\usepackage{mathabx}
\usepackage{mathrsfs}
\usepackage{listings}

\usetikzlibrary{matrix,chains,positioning,decorations.pathreplacing,arrows}
\usetikzlibrary{chains,scopes}

\usepackage{placeins}

\usetikzlibrary{arrows.meta} 
\usepackage[outline]{contour} 
\contourlength{1.4pt}

\usepackage{svg}
\usepackage{xcolor}

\newcommand{\printfnsymbol}[1]{%
	\textsuperscript{\@fnsymbol{#1}}%
} 
\makeatother

\usepackage{accvabbrv}

\usepackage{graphicx}
\usepackage{booktabs}

\usepackage[accsupp]{axessibility}  

\usepackage{hyperref}

\usepackage{orcidlink}

\begin{document}

\title{CNN Mixture-of-Depths} 

\titlerunning{CNN MoD}

\author{Rinor Cakaj\inst{1,2}\and
Jens Mehnert\inst{1} \and
Bin Yang\inst{2}}

\authorrunning{Cakaj, Mehnert, Yang.}

\institute{Robert Bosch GmbH, Daimlerstrasse 6, 71229 Leonberg, Germany\\\email{\{Rinor.Cakaj,JensEricMarkus.Mehnert\}@de.bosch.com}\and
	University of Stuttgart, Pfaffenwaldring 47, 70550 Stuttgart, Germany\\\email{bin.yang@iss.uni-stuttgart.de}}

\maketitle

\begin{abstract}
	
We introduce Mixture-of-Depths (MoD) for Convolutional Neural Networks (CNNs), a novel approach that enhances the computational efficiency of CNNs by selectively processing channels based on their relevance to the current prediction. This method optimizes computational resources by dynamically selecting key channels in feature maps for focused processing within the convolutional blocks (Conv-Blocks), while skipping less relevant channels. Unlike conditional computation methods that require dynamic computation graphs, CNN MoD uses a static computation graph with fixed tensor sizes which improve hardware efficiency. It speeds up the training and inference processes without the need for customized CUDA kernels, unique loss functions, or fine-tuning.
CNN MoD either matches the performance of traditional CNNs with reduced inference times, GMACs, and parameters, or exceeds their performance while maintaining similar inference times, GMACs, and parameters. For example, on ImageNet, ResNet86-MoD exceeds the performance of the standard ResNet50 by 0.45\% with a 6\% speedup on CPU and 5\% on GPU. Moreover, ResNet75-MoD achieves the same performance as ResNet50 with a 25\% speedup on CPU and 15\% on GPU.

\keywords{CNN, Mixture-of-Depths, Computational Efficiency, Inference Speed}

\end{abstract}

\section{Introduction}
\label{sec:intro}

Over recent years, convolutional neural networks (CNNs) have demonstrated significant advancements in a variety of computer vision applications, such as image recognition \cite{2016_He_CONF, 2015_Szegedy_CONF}, object detection \cite{2016_Redmon_CONF, 2015_Ren_CONF}, and image segmentation \cite{2015_Long_CONF, 2016_Yu_CONF}. Despite their remarkable performance, CNNs often require substantial computational power and extensive memory usage, posing considerable challenges when deploying advanced models on devices with limited computational resources \cite{2021_Li}.

To adress these challenges, pruning techniques are widely utilized to reduce the model size and computational demands of CNNs by removing redundant weights or filters according to established criteria \cite{2019_He_CONF, 2017_Li_CONF, 2017_Luo_CONF, 2018_Zhuang_CONF, 2018_Yu_CONF, 2018_He_CONF, 2017_He_CONF, 2022_Wimmer_CONF}. However, these methods uniformly process all inputs, failing to adjust for the varying complexities of different inputs, which can lead to performance decreases \cite{2021_Li}.

An alternative approach is dynamic computing, or conditional computation, which adapts computational resources to the complexity of inputs to enhance efficiency \cite{2021_Li, 2018_Veit_CONF, 2020_Verelst_CONF}. However, the integration of these methods into hardware is challenging due to their reliance on dynamic computation graphs, which are often incompatible with systems optimized for static computation workflows \cite{2020_Verelst_CONF, 2016_Graves}. For instance, Wu et al. \cite{2018_Wu_CONF} report increased processing times when layers are conditionally executed via a separate policy network. Despite theoretical reductions in computational demands, practical gains on hardware like GPUs or FPGAs are limited, as non-uniform tasks can interfere with the efficiencies of standard convolution operations \cite{2017_Sze, 2016_Lavin_CONF}. Moreover, Ma et al. \cite{2018_Ma_CONF} show that floating point operations (FLOPS) are insufficient for estimating inference speed, as they often exclude element-wise operations like activation functions, summations, and pooling.

\begin{figure*}[tb]
	\includegraphics[width=\linewidth]{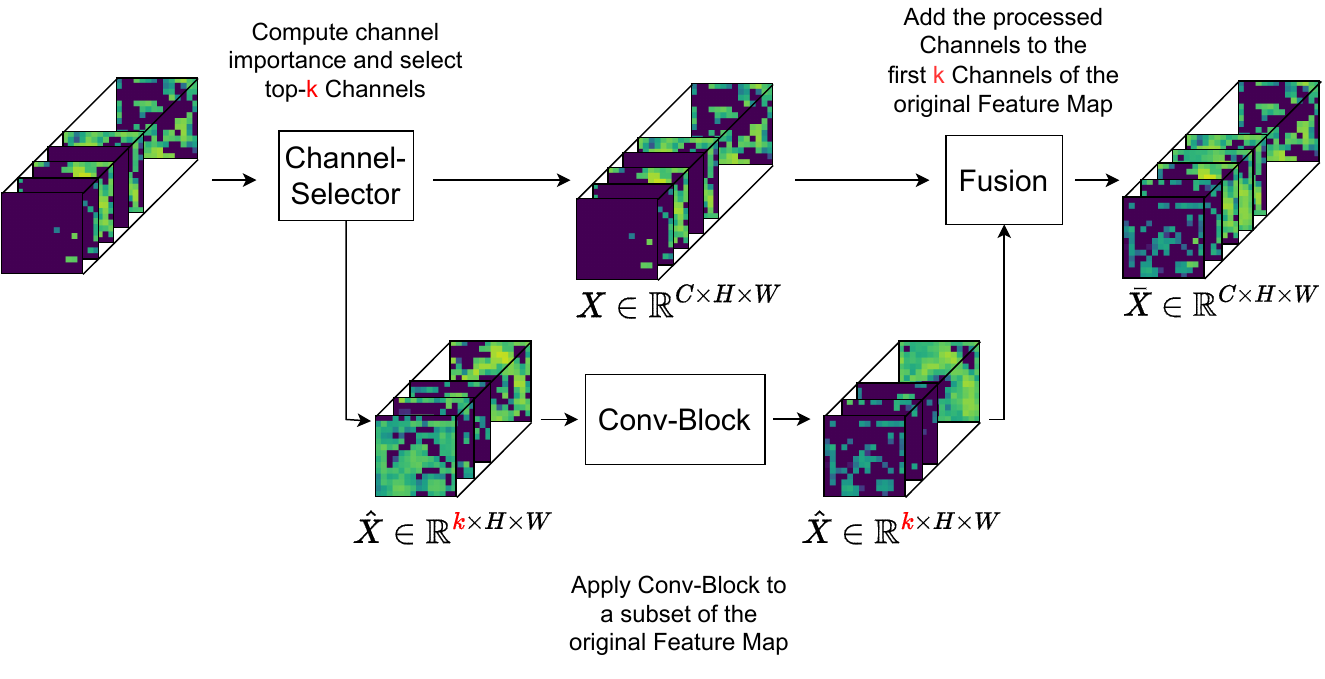}
	\caption{Illustration of the CNN MoD mechanism, which starts with the Channel Selector module. This module computes the importance scores of each channel in the input feature map, \(X \in \mathbb{R}^{C \times H \times W}\), and selects the top-\(k\) channels for focussed processing in the Conv-Block. These selected channels are then processed by a Conv-Block designed to operate on a reduced dimension, \( \hat{X} \in \mathbb{R}^{k \times H \times W} \), enhancing computational efficiency. The processed channels are added to the first \(k\) channels of the original feature map through a fusion operation, instead of being added back to their original positions. The resulting feature map is denoted by \( \bar{X} \). This selective reintegration of refined channels with the unprocessed channels helps to preserve the dimensions of the original feature map (\(X \in \mathbb{R}^{C \times H \times W}\)).
	}
	\label{fig:CNN_MoD_illustration}
\end{figure*}

To combine the performance benefits of dynamic computing with the operational efficiency of static computation graphs, we present the CNN Mixture-of-Depths (MoD), inspired by the Mixture-of-Depths approach for Transformers \cite{2024_Raposo}. The fundamental principle of CNN MoD is that only a selected subset of channels within the feature maps is essential for effective convolutional processing at given layers in the network. By focusing on these essential channels, CNN MoD boosts computational efficiency without reducing network performance.

The MoD mechanism, shown in Figure \ref{fig:CNN_MoD_illustration}, enhances CNN feature map processing by dynamically selecting the most important channels for focused computation within Conv-Blocks. MoD begins with the Channel Selector, illustrated in Figure \ref{fig:CNN_MoD_Channel_Selector}, which is designed similarly to the Squeeze-and-Remember block \cite{2018_Hu_CONF}. It evaluates the importance of each channel within the input feature map \( X \in \mathbb{R}^{C \times H \times W} \). Adaptive average pooling first reduces each channel to \( \tilde{X} \in \mathbb{R}^{C \times 1 \times 1} \), which is then processed by a two-layer fully connected neural network with a bottleneck design. A sigmoid activation function generates scores indicating the importance of each channel.

Following this, the Channel Selector uses a top-$k$ selection mechanism to identify the \( k \) most crucial channels based on the computed importance scores.
These channels are sent to the Conv-Block, designed to operate on the reduced dimension \( \hat{X} \in \mathbb{R}^{k \times H \times W} \), thus improving computational efficiency. To ensure that the gradients from the processed output \( \hat{X} \) effectively optimize the channel selection process, the processed channels are scaled by their respective importance scores. This allows the gradients to flow back to the Channel Selector, enabling the learning of channel importance throughout training.

The final step involves a fusion operation, in which the processed channels are added to the first \( k \) channels of the original feature map. This fusion not only preserves the original dimensions of the feature map but also enhances feature representation by combining processed and unprocessed channels.

The reduction in the number of processed channels within Conv-Blocks is controlled by a hyperparameter $c \geq 1$, where \( k = \lfloor \frac{C}{c} \rfloor \) defines the number of channels to process. Here, \(C\) is the total number of input channels in the Conv-Block. For example, in a typical ResNet \cite{2016_He_CONF} architecture, a bottleneck block that usually processes 1024 channels will only process 16 channels (\(k = 16\)) when \(c = 64\). This selective processing significantly reduces the computational load by focusing on a smaller subset of channels. Additionally, the sizes of the kernels within the Conv-Blocks are adjusted to match the reduced number of input channels, as detailed in Section \ref{subsec:conv_path}.

Empirical evaluations indicate that the optimal integration of MoD within the CNN architecture involves alternating them with standard Conv-Blocks. In architectures like ResNets \cite{2016_He_CONF} or MobileNetV2 \cite{2018_Sandler_CONF}, Conv-Blocks are organized into modules containing multiple Conv-Blocks of the same type (i.e., with the same number of output channels). Each module begins with a standard block, and is then followed by a MoD Block. This alternating arrangement is based on the principles of the Mixture-of-Depths for Transformers \cite{2024_Raposo}. This does not imply that additional MoD Blocks are added to the existing sequence, but rather every second Conv-Block in the original architecture is replaced by a MoD Block, ensuring the overall depth of the architecture remains unchanged.

CNN MoD achieves performance comparable to traditional CNNs but with reduced inference times, GMACs, and parameters, or it surpasses them while maintaining similar inference times, GMACs, and parameters. On the ImageNet dataset \cite{2015_Russakovsky}, our ResNet75-MoD matches the accuracy of the standard ResNet50 \cite{2016_He_CONF} and provides speed-ups of 15\% on GPU and 25\% on CPU. Similar results can also be achieved in semantic segmentation and object detection tasks, as detailed in Section \ref{experiments}.

\section{Related Work}

This section reviews two strategies for enhancing CNN computational efficiency: static pruning and dynamic computing. At the end, we highlight the advantages of our CNN MoD approach, which combines the benefits of both strategies, offering improvements over traditional methods.

\subsection{Static Pruning}

Static pruning techniques aim to reduce the computational burden of CNNs by eliminating redundant model parameters. Early work in this area primarily targeted weight pruning, which selectively removes individual weights that have little effect on the model's output \cite{2016_Guo_CONF, 2015_Han_CONF, 2021_Wimmer_CONF}. However, these methods tend to create irregular sparsity patterns that do not align well with hardware optimizations \cite{2021_Li}. More recently, structured pruning approaches like channel pruning have gained prominence. These methods offer a more hardware-compatible form of sparsity by discarding entire channels based on their assessed importance \cite{2019_He_CONF, 2020_Lin_CONF}. For example, FPGM, or Filter Pruning via Geometric Median, identifies and removes filters that are closest to the geometric median within a layer, targeting those considered less crucial \cite{2019_He_CONF}. Similarly, HRank evaluates filters based on the rank of their generated feature maps, pruning those that contribute the least to the output's information content \cite{2020_Lin_CONF}. 

Despite their benefits, these static pruning methods permanently remove computations, potentially reducing the model's capacity to represent complex features. This permanent reduction can be particularly limiting for complex images that may require more detailed processing, suggesting that a one-size-fits-all approach to pruning might be suboptimal for diverse real-world applications. Additionally, these methods often require fine-tuning or specialized loss functions, further complicating their implementation.

\subsection{Dynamic Computing}

Dynamic computing, or conditional computation, dynamically adjusts computational resources according to the complexity of the input, potentially maintaining high model accuracy while reducing computation. Architectures like BranchyNet \cite{2016_Teerapittayanon_CONF} and MSDNet \cite{2018_Huang_CONF} implement early exits for simpler inputs to decrease average computational load. Furthermore, policy-driven methods such as BlockDrop \cite{2018_Wu_CONF} and GaterNet \cite{2019_Chen_CONF} dynamically decide which network blocks to execute using a policy network, which adapts in real-time to the input.

Further innovations in dynamic computing utilize gating functions to enable selective processing of Conv-Blocks. SkipNet \cite{2018_Wang_CONF} and ConvNet-AIG \cite{2018_Veit_CONF} dynamically skip the processing of whole Conv-Blocks based on the observation that individual blocks can be removed without interfering with other blocks in the ResNet \cite{2016_He_CONF}. Some methods \cite{2017_Dong_CONF, 2020_Verelst_CONF,  2020_Xie_CONF} exploit spatial sparsity to reduce the computations, e.g. Verelst et al. \cite{2020_Verelst_CONF} learn a pixel-wise execution mask for each block and only calculate on those specified locations. Dynamic Dual Gating \cite{2021_Li} introduces another layer of complexity by identifying informative features along spatial and channel dimensions, allowing the model to skip unimportant regions and channels dynamically during inference. 

Many works on conditional execution primarily highlight reductions in theoretical complexity. However, in practical settings, the application of these dynamic methods often confronts hardware limitations due to their dependency on dynamic computation graphs. This reliance can lead to inefficiencies and extended execution times on hardware systems that are optimized for static computations \cite{2016_Graves, 2018_Wu_CONF, 2017_Sze,2016_Lavin_CONF}.

Recently, Raposo et al. \cite{2024_Raposo} introduced the Mixture-of-Depths for Transformers, a method that selectively processes tokens within Transformer blocks to effectively reduce computational load. This innovative approach inspired us for our CNN MoD approach.

\subsection{CNN MoD: Combining Static and Dynamic Advantages}

The CNN MoD approach combines the benefits of static pruning and dynamic computing within a unified framework. Below are the key advantages of our approach:

\begin{itemize}
	\item \textbf{Static Computation Graph:} CNN MoD retains a static computation graph, which enhances both training and inference time.
	\item \textbf{No Custom Requirements:} Does not require customized CUDA kernels, additional loss functions, or fine-tuning.
	\item \textbf{Dynamic Resource Allocation:} Dynamically allocates computational resources to channels based on their importance, effectively optimizing training and inference speeds.
\end{itemize}

\section{Method: CNN Mixture-of-Depths}
\label{method}

The term ``Mixture-of-Depths'' in our approach refers to the selective processing strategy where not every channel is processed in every convolutional block within the same module, resulting in varied processing depths. This strategy allows for some channels to be processed more frequently than others within the same module, thereby optimizing computational resources and enhancing feature extraction efficiency.
This section presents the CNN Mixture-of-Depths approach. It consists of three main components:

\begin{enumerate}
	\item \textbf{Channel Selector:} This component selects the top-$k$ most important channels from the input feature map based on their relevance to the current prediction. This selection helps to focus computational resources effectively by processing only those channels that are crucial for the current task.
	
	\item \textbf{Convolutional Block:} The selected channels are then processed in a Conv-Block. This block can be adapted from existing architectures such as ResNets \cite{2016_He_CONF} or ConvNext \cite{2022_Liu_CONF} and is designed to enhance the features of the selected channels. To ensure that the Channel Selector is optimized during training, the processed feature maps are multiplied by their importance scores at the end of this block.
	
	\item \textbf{Fusion Operation:} After processing in the Conv-Block, the enhanced channels are integrated back into the original feature map. This is done by adding the processed channels to the first $k$ channels of the feature map, which is then passed on to the subsequent layer without additional adjustments. This step ensures that the processed features are preserved and that the feature map maintains its original dimensions for further processing.
\end{enumerate}

\subsection{Channel Selector}
\label{subsec:channel_selector}

\begin{figure*}[tb]
	\centering
	\includegraphics[width=\linewidth]{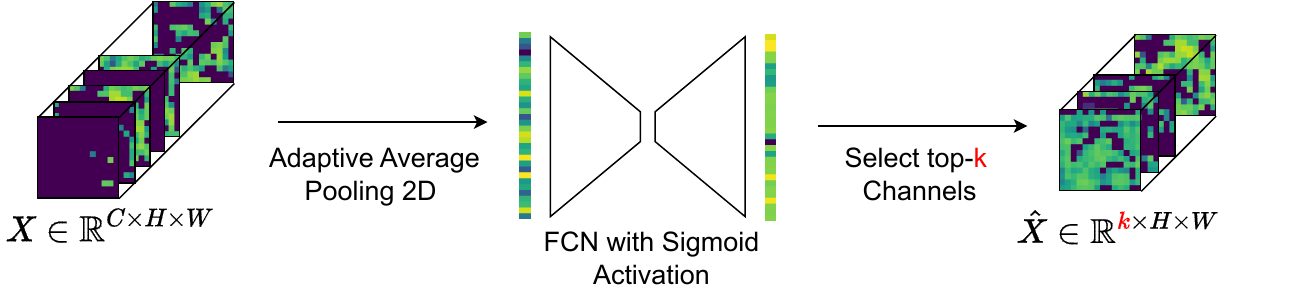}
	\caption{Illustration of the Channel Selection Process in MoD. The process begins with an input tensor $X \in \mathbb{R}^{C \times H \times W}$, which undergoes adaptive average pooling to reduce spatial dimensions to $1 \times 1$, preserving channel information. The pooled output is processed through a two-layer fully connected network with a reduction factor ($r=16$), followed by a sigmoid activation to generate channel-wise scores. These scores are used to select the top-$k$ channels. This forms a subset of the original tensor with reduced channel dimension but original spatial dimensions.}
	\label{fig:CNN_MoD_Channel_Selector}
\end{figure*}

Figure \ref{fig:CNN_MoD_Channel_Selector} shows how the Channel Selector dynamically selects the most significant channels for processing in the Conv-Block. It operates in two main stages:

\begin{enumerate}
	\item \textbf{Adaptive Channel Importance Computation:} Initially, the Channel Selector compresses the input feature map \( X \in \mathbb{R}^{C \times H \times W} \) using adaptive average pooling, reducing its dimension to \( \tilde{X} \in \mathbb{R}^{C \times 1 \times 1} \). This compressed feature map is processed through a two-layer fully connected network with a bottleneck design, set to a ratio \( r = 16 \), and concludes with a sigmoid activation to create a score vector \( \mathbf{s} \in \mathbb{R}^C \). Each element of \( \mathbf{s} \) quantifies the importance of the corresponding channel.
	
	\item \textbf{Top-k Channel Selection and Routing:} Utilizing the importance scores \( \mathbf{s} \), the Channel Selector selects the top-$k$ channels. These selected channels are routed to the Conv-Block for processing. The original feature map \( X \) is routed around the Conv-Block to the fusion process described in Section \ref{subsec:fusion-mechanism}.
	
\end{enumerate}

This selection process allows the Channel Selector to efficiently manage computational resources while maintaining a fixed computational graph, making it possible to dynamically choose which channels to process.

\subsection{Channel Processing Dynamics}
\label{subsec:conv_path}

After the Channel Selector selects the top-$k$ channels, they are processed within the Conv-Block. This block is adaptable from architectures  such as ResNets \cite{2016_He_CONF} or ConvNext \cite{2022_Liu_CONF}, and is designed to process a reduced number of channels.

The number of channels \( k \) processed in each Conv-Block is determined by the formula \( k = \lfloor \frac{C}{c} \rfloor \), where \( C \) represents the total input channels of the block, and \( c \) is a hyperparameter determining the extent of channel reduction. For instance, in a standard ResNet \cite{2016_He_CONF} bottleneck block that typically processes 1024 channels, setting \( c = 64 \) reduces the processing to only 16 channels (\( k = 16 \)). Unlike the standard bottleneck flow where channels transition from 1024 to 256 and back to 1024, in a MoD bottleneck block, the channel dimensions are significantly reduced from 16 to 4 and then back to 16, focusing computational efforts on the most essential features and enhancing efficiency. In our empirical investigations, we found that the hyperparameter \(c\) should be set to the maximal number of input channels in the first Conv-Block and remain the same in every MoD Block throughout the CNN. For example, \(c = 64\) for ResNet \cite{2016_He_CONF} (see Appendix \ref{channel_param_effect}) and \(c = 16\) for MobileNetV2 \cite{2018_Sandler_CONF}.

The final step in the Conv-Block involves multiplying the processed channels with the importance scores obtained from the Adaptive Channel Importance Computation. This step ensures that gradients are effectively propagated back to the Channel Selector during training, which is needed for optimizing the selection mechanism.

\subsection{Fusion Mechanism}
\label{subsec:fusion-mechanism}

The Fusion Mechanism combines processed and unprocessed channels to maintain the dimensionality required for subsequent convolutional operations. This approach ensures that the remaining non-selected channels, which contain useful features for later stages, are preserved, thereby maintaining a comprehensive feature set necessary for the model's performance.

Consider the original input feature map \( X \) with dimensions \( \mathbb{R}^{C \times H \times W} \), where \( C \) represents the total number of channels, and \( H \) and \( W \) denote the height and width of the feature map, respectively. The processed feature map \( \hat{X} \), resulting from the adapted Conv-Block applied to the selected top-\( k \) channels, has dimensions \( \mathbb{R}^{k \times H \times W} \). To integrate \( \hat{X} \) with \( X \), the fusion operation is performed as follows:
\begin{align}
	\bar{X}[:, :k, :] &= \hat{X} + X[:, :k, :], \tag{1} \\
	\bar{X}[:, k:, :] &= X[:, k:, :]. \tag{2}
\end{align}
where (1) adds processed features to the first \( k \) channels of \( X \), and (2) maintains the remaining unprocessed channels. This formula confirms that \( \bar{X} \), the feature map after fusion, has the same number of channels \( C \) as the original input \( X \), preserving the dimensions needed for subsequent layers. 

In our experiments, we tested various strategies for integrating the processed channels back into the feature map \( X \). These strategies included adding the processed channels back to their original positions. However, empirical results did not show any improvements (see Appendix \ref{channel_param_effect}). It seems beneficial for the network's performance to consistently use the same positions within the feature maps for processed information. This observation is confirmed by experiments where the processed channels were added to the last \(k\) channels of the feature map \(X\), yielding results comparable to those achieved when added to the first \(k\) channels (see Appendix \ref{lastk_results_imagenet}).

\subsection{Integration in CNN Architecture}
\label{subsec:integration-resnet}

MoD can be integrated into various CNN architectures, such as ResNets \cite{2016_He_CONF}, ConvNext \cite{2022_Liu_CONF}, VGG \cite{2015_Simonyan_CONF}, and MobileNetV2 \cite{2018_Sandler_CONF}. These architectures are organized into modules containing multiple Conv-Blocks of the same type (i.e., with the same number of output channels). Through our experiments, we found out that alternating MoD Blocks with standard Conv-Blocks in each module is the most effective integration method. We also explored using MoD Blocks in every block and selectively within specific modules, however, the alternating strategy proved to be the most effective approach. It is important to note that MoD Blocks replace every second Conv-Block, maintaining the original architecture's depth (e.g., 50 layers in a ResNet50 \cite{2016_He_CONF}). Each module starts with a standard block, such as a BasicBlock \cite{2016_He_CONF}, followed by an MoD Block. This alternating pattern indicates that the network can handle substantial capacity reductions, provided that there are regular intervals of full-capacity convolutions. Furthermore, this method ensures that MoD Blocks do not interfere with the spatial dimension-reducing convolutions that typically occur in the first block of each module.

\section{Experiments}
\label{experiments}

\begin{figure}[tb]
	\centering
	\begin{subfigure}[b]{0.49\textwidth}
		\centering
		\includegraphics[width=\textwidth]{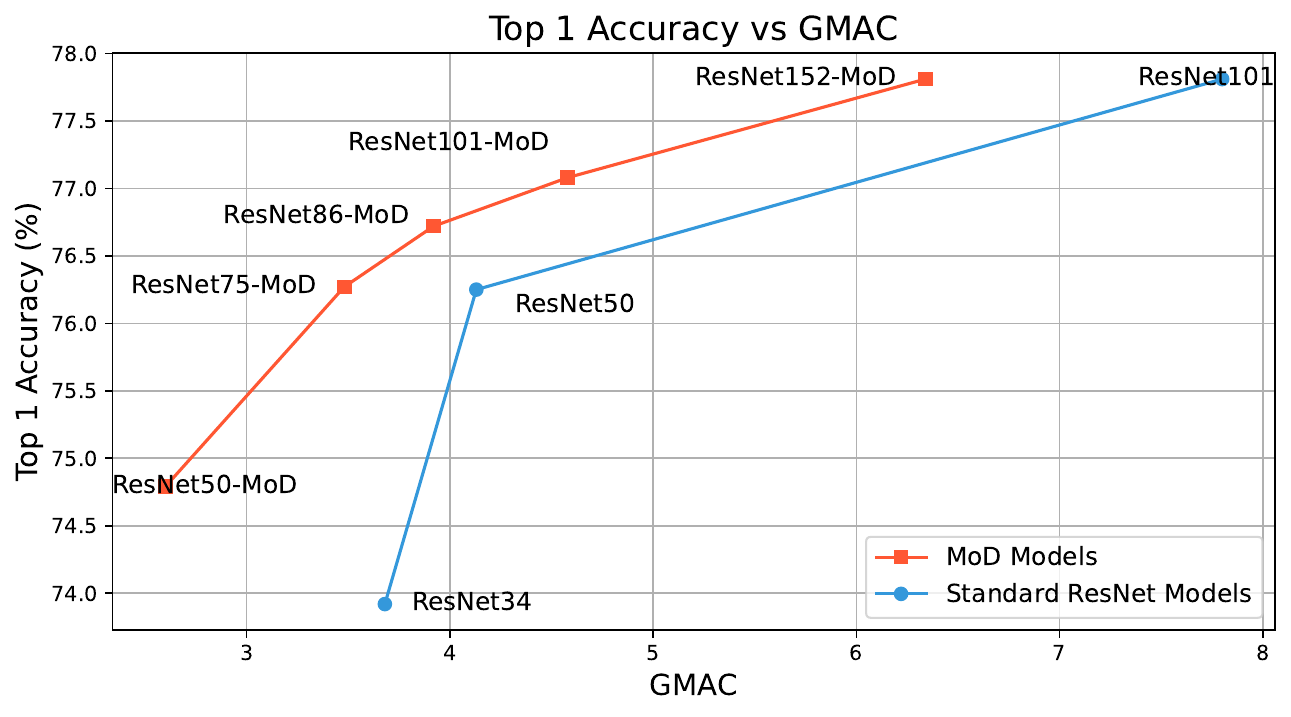}
		\captionsetup{justification=centering}
		\caption{Top-1 Accuracy vs GMAC}
		\label{fig:gmac}
	\end{subfigure}
	\hfill
	\begin{subfigure}[b]{0.49\textwidth}
		\centering
		\includegraphics[width=\textwidth]{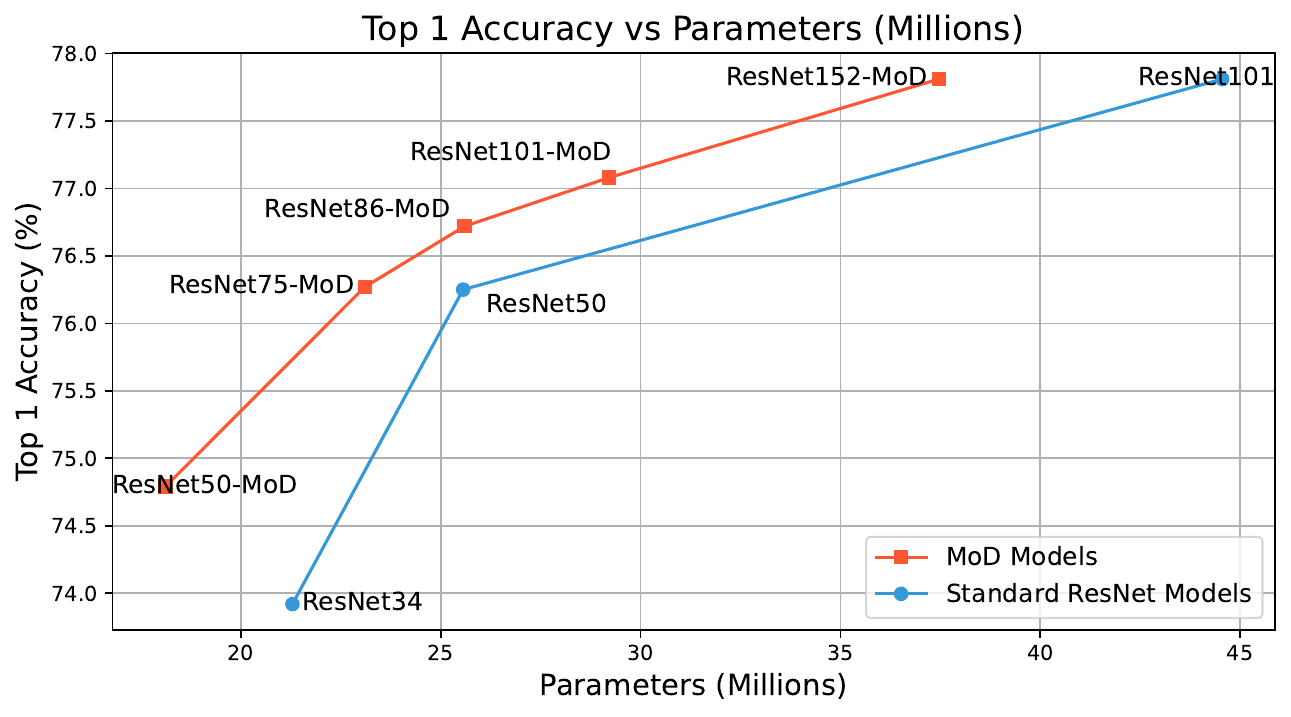}
		\captionsetup{justification=centering}
		\caption{Top-1 Accuracy vs Parameters}
		\label{fig:parameters}
	\end{subfigure}
	\vspace{1em} 
	\begin{subfigure}[b]{0.49\textwidth}
		\centering
		\includegraphics[width=\textwidth]{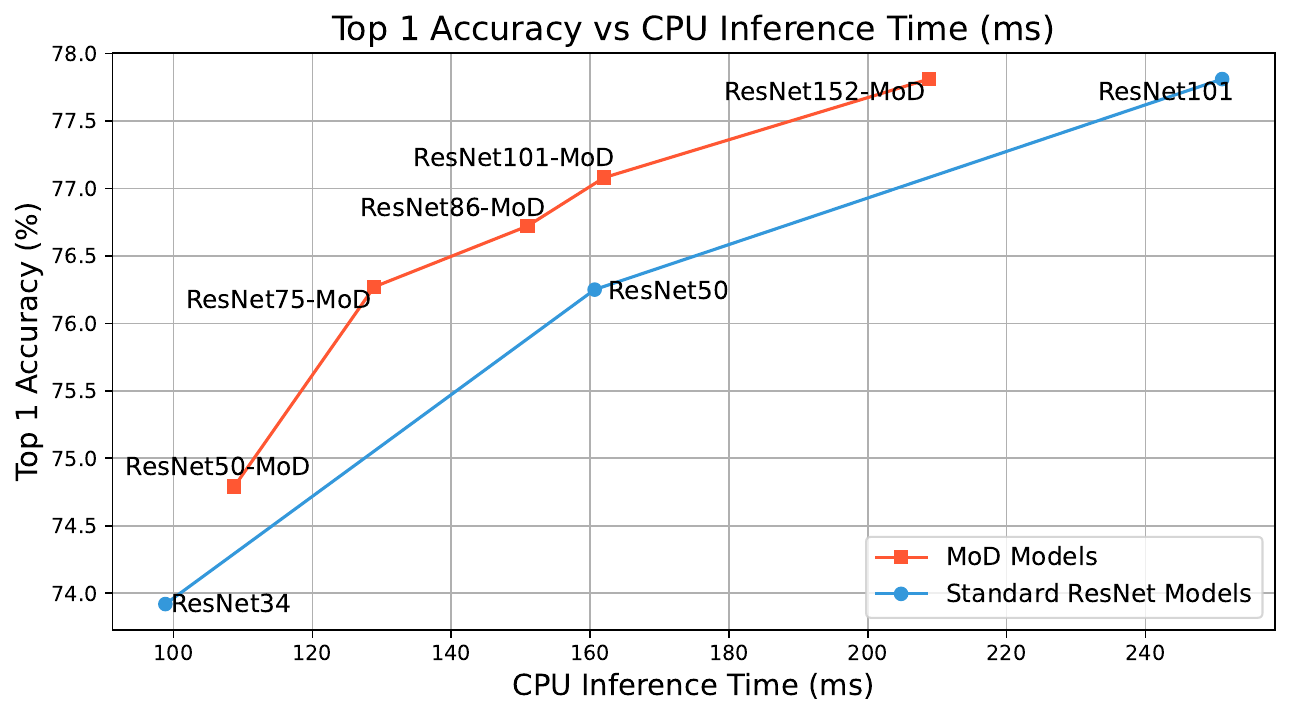}
		\captionsetup{justification=centering}
		\caption{Top-1 Accuracy vs CPU Inference Time}
		\label{fig:cpu_inference}
	\end{subfigure}
	\hfill 
	\begin{subfigure}[b]{0.49\textwidth}
		\centering
		\includegraphics[width=\textwidth]{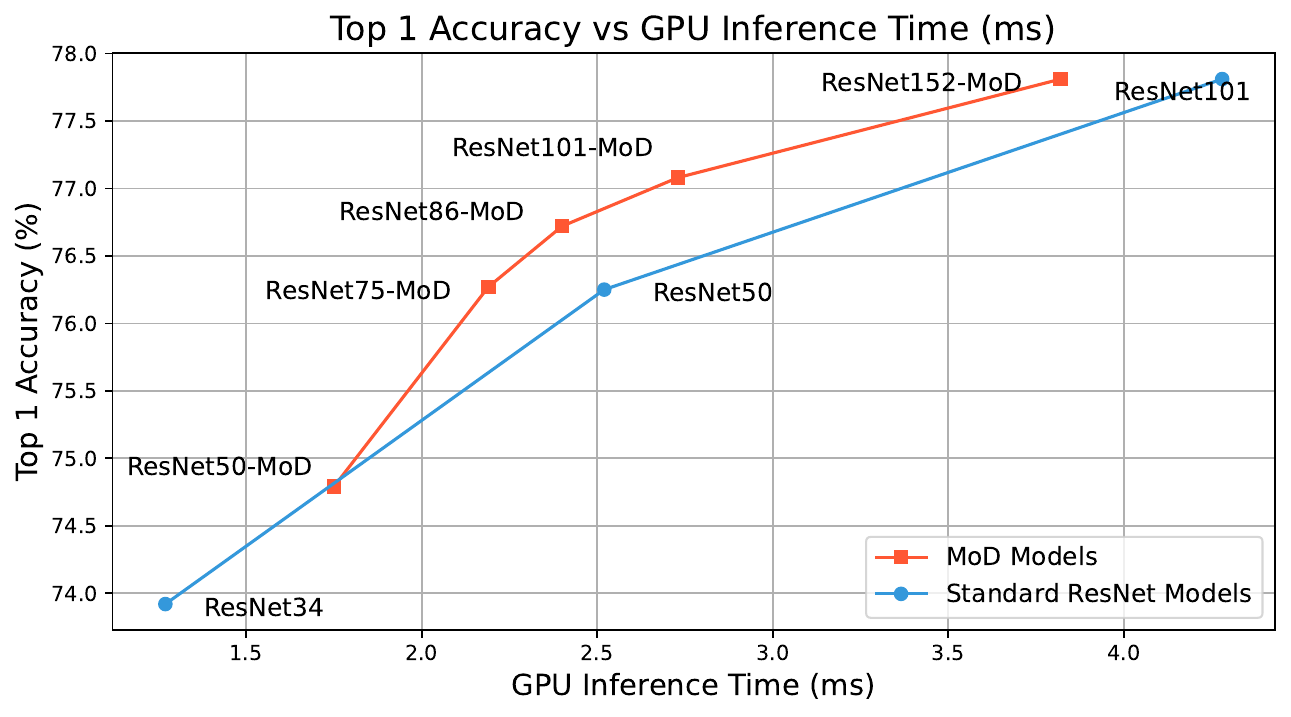}
		\captionsetup{justification=centering}
		\caption{Top-1 Accuracy vs GPU Inference Time}
		\label{fig:gpu_inference}
	\end{subfigure}

	\caption{ResNet MoD models outperform standard ResNets under similar computational constraints, as shown across four panels. Panel (a) shows the higher accuracy per GMAC, highlighting better computational efficiency, while panel (b) illustrates the improved parameter efficiency. Panels (c) and (d) demonstrate ResNet MoD's superior top-1 accuracy with comparable or faster inference times on CPU and GPU.}
	\label{fig:combined}
\end{figure}

This section evaluates the CNN Mixture-of-Depths on various tasks, including supervised image recognition on CIFAR-10/100 \cite{2009_Krizhevsky} and ImageNet \cite{2015_Russakovsky}, semantic segmentation on Cityscapes \cite{2016_Cordts_CONF}, and object detection on Pascal VOC \cite{2010_Everingham}. Results for CIFAR-10/100 are detailed in Appendix \ref{exp_CIFAR}.

Performance benchmarks were carried out on both GPU and CPU platforms to evaluate the running times. These measurements were taken using the Timer and Compare functions from the \texttt{torch.utils.benchmark} package, with tests performed on an Intel(R) Core(TM) i7-10850H CPU @ 2.70GHz and a NVIDIA Quadro RTX 3000 GPU.

\subsection{Image Recognition on ImageNet} \label{exp_imagenet}

We evaluated the CNN MoD on the ImageNet dataset \cite{2015_Russakovsky}, containing 1.2 million training images, 50,000 validation images, and 150,000 test images across 1,000 categories. The experiments were performed using three different random seeds to ensure the robustness and reproducibility of the results. Implementation details and further results are provided in Appendix \ref{appendix_imagenet}.

Table~\ref{imagenet_table} presents the comparative results for various network configurations, highlighting our approach against advanced pruning and dynamic computing strategies. This includes comparisons with established methods such as DGNet \cite{2021_Li}, Batch-Shaping \cite{2020_Bejnordi_CONF}, ConvNet-AIG \cite{2018_Veit_CONF}, HRANK \cite{2020_Lin_CONF}, FPGM \cite{2019_He_CONF}, and DynConv \cite{2020_Verelst_CONF}. We report for all models the top-1 classification accuracy on the validation set, along with attributes such as no requirement for fine-tuning (No FT), simple training mechanisms (Simple Train.), the availability of a GPU implementation (GPU Impl.), a static computational graph (Static Graph), and enhanced training speed (Faster Train.). Here, ''Simple Training`` indicates that no customized loss functions or complex hyperparameter tuning is required, and ''Faster Training`` is determined by measuring the relative training time compared to standard models.

The MoD Blocks reduce the computational load in every second block, allowing the construction of deeper architectures with the computational costs comparable to less deep standard models. Specifically, in the ResNet86-MoD, the number of Conv-Blocks in the third layer is 18, compared to just 6 blocks in the standard ResNet50, while maintaining nearly the same GMACs, parameters, and inference times. This expansion is enabled by the MoD Blocks that process fewer channels. Similarly, the ResNet75-MoD increases the number of Conv-Blocks in the third layer to 14. For further details, see Appendix \ref{architecture_details_resnet_imagenet}.

As described in Section \ref{subsec:conv_path}, the value of \(c\) in all MoD Blocks is set to the maximum number of input channels in the first Conv-Block. For instance, \(c = 64\) for ResNet \cite{2016_He_CONF} and \(c = 16\) for MobileNetV2 \cite{2018_Sandler_CONF}.

The MoD models achieve performances comparable to traditional CNNs but with reduced inference times or they surpass them while maintaining similar inference times. For instance, our ResNet75-MoD matches the performance of the standard ResNet50 and achieves an 15\% speed-up on GPU and 25\% on CPU. Similarly, the ResNet50-MoD model significantly enhances processing speeds, achieving a 27\% improvement on CPU and 48\% on GPU, with a slight impact on accuracy. Furthermore, the ResNet86-MoD model improves accuracy by 0.45\% with a speed-up of 6\% on CPU and 5\% on GPU. A test variant, MoD50 rand.\ using randomized channel selection, shows a performance decline of about 0.8\%. This highlights the significance of strategic channel selection in our MoD approach. Figure \ref{fig:combined} presents a comparative performance analysis of ResNet MoD models against standard ResNets for different computational metrics.

Dynamic approaches like DGNet \cite{2021_Li} and ConvNet-AIG \cite{2018_Veit_CONF}, which lack data on real-world inference speed-up, show simulated GMAC reductions only, indicating potential discrepancies in practical applications. 

The results presented in Table \ref{imagenet_mobilenet_table} demonstrate the effectiveness of applying the MoD approach to MobileNetV2 architectures. The MobileNetV2-MoD-L model, with a deeper configuration (see Appendix \ref{architecture_details_mobilenetv2_imagenet}), maintains nearly the same top-1 accuracy as the standard MobileNetV2, while achieving a 11\% speed-up on CPU and 10\% on GPU. The standard MobileNetV2-MoD, despite a slight decrease in accuracy, offers a significant speed-up of 43\% on CPU and 39\% on GPU, illustrating the benefits of the MoD approach in balancing performance with computational efficiency.

\begin{table}[tb]
	\centering
	\setlength{\tabcolsep}{3pt}
	\renewcommand{\arraystretch}{1.3}
	\caption{This table presents the MoD models against other dynamic pruning and computational efficiency strategies. The table evaluates each method based on top-1 accuracy (Acc \%), computational complexity (GMAC), model size (Params, in millions - M) and speed-up for CPU and GPU (Speed-up CPU, Speed-up GPU). Models abbreviated as ``R50'' refer to ResNet50 architectures. Notably, only MoD models incorporate all the attributes such as No Fine-Tuning (No FT), Simple Training (Simple Train.), GPU Implementation (GPU Impl.), a Static Computational Graph (Static Graph), and Enhanced Training Speed (Faster Train.).} \label{imagenet_table}
	\begin{threeparttable}
		\begin{adjustbox}{max width=\textwidth}
			\begin{tabular}{lcccccccccccc}
				\toprule
				\textbf{Model} & \textbf{No} & \textbf{Simple} & \textbf{GPU} & \textbf{Static} & \textbf{Faster} & \textbf{Top-1} & \textbf{GMAC} & \textbf{Params} & \multicolumn{2}{c}{\textbf{Inference (ms)}} & \multicolumn{2}{c}{\textbf{Speed-up}} \\
				\cmidrule(lr){10-11} \cmidrule(lr){12-13}
				& \textbf{FT} & \textbf{Train.} & \textbf{Impl.} & \textbf{Graph} & \textbf{Train.} & \textbf{Acc (\%)} &  & \textbf{(M)} & \textbf{CPU} & \textbf{GPU} & \textbf{CPU} & \textbf{GPU} \\
				\midrule
				\rowcolor{gray!20} \textbf{ResNet152-MoD} & \cmark & \cmark & \cmark & \cmark & \cmark & 77.81 & 6.34 & 37.46 & 208.80 & 3.82 & 1.20 & 1.12 \\
				ResNet101 &  &  &  &  &  & 77.81 & 7.80 & 44.55 & 251.05 & 4.28 & 1.00 & 1.00 \\
				\rowcolor{gray!20} \textbf{ResNet101-MoD} & \cmark & \cmark & \cmark & \cmark & \cmark & 77.08 & 4.58 & 29.21 & 162.03 & 2.73 & 1.55 & 1.57 \\
				\midrule
				\rowcolor{gray!20} \textbf{ResNet86-MoD} & \cmark & \cmark & \cmark & \cmark & \cmark & 76.72 & 3.92 & 25.60 & 150.96 & 2.40 & 1.06 & 1.05 \\
				R50-DGNet \cite{2021_Li} & \cmark & \xmark & \xmark & \xmark & \xmark & 76.41 & 1.65\tnote{*} & 29.34 & {---}\tnote{1} & {---}\tnote{1} & {---} & {---} \\
				\rowcolor{gray!20} \textbf{ResNet75-MoD} & \cmark & \cmark & \cmark & \cmark & \cmark & 76.27 & 3.48 & 23.10 & 128.90 & 2.19 & 1.25 & 1.15 \\
				ResNet50 &  &  &  &  &  & 76.25 & 4.13 & 25.56 & 160.66 & 2.52 & 1.00 & 1.00 \\
				R50-Batch-Shaping \cite{2020_Bejnordi_CONF} & \cmark & \xmark & \xmark & \xmark & \xmark & 75.70 & 2.07\tnote{*} & 15.31 & {---} & {---} & 1.31\tnote{2} & {---} \\
				R50-ConvNet-AIG \cite{2018_Veit_CONF} & \cmark & \xmark & \xmark & \xmark & \xmark & 75.45 & 2.59\tnote{*} & 26.56 & {---} & {---} & 1.29\tnote{2} & {---} \\
				R50-HRANK \cite{2020_Lin_CONF} & \xmark & \xmark & \cmark & \cmark & \xmark & 74.98 & 2.30 & 16.15 & {---}\tnote{3} & {---}\tnote{3} & {---} & {---} \\
				\rowcolor{gray!20} \textbf{ResNet50-MoD} & \cmark & \cmark & \cmark & \cmark & \cmark & 74.79 & 2.60 & 18.11 & 108.74 & 1.75 & 1.48 & 1.44 \\
				R50-FPGM \cite{2019_He_CONF} & \xmark & \xmark & \cmark & \cmark & \xmark & 74.83 & 1.91 & {---} & {---} & {---} & {---} & 1.61\tnote{4} \\
				R50-DynConv \cite{2020_Verelst_CONF} & \cmark & \xmark & \cmark & \xmark & \xmark & 74.40 & 2.25 & 25.56 & {---} & {---} & {---}  & {---}\tnote{5}  \\
				R50-FPGM \cite{2019_He_CONF} & \cmark & \xmark & \cmark & \cmark & \xmark & 74.13 & 1.91 & {---} & {---} & {---} & {---} & 1.61\tnote{4} \\
				\rowcolor{gray!20} \textbf{MoD50 rand.} & \cmark & \cmark & \cmark & \cmark & \cmark & 74.02 & 2.59 & 17.11 & 100.06 & 1.43 & 1.77 & 1.43 \\
				\bottomrule
			\end{tabular}
		\end{adjustbox}
		\begin{tablenotes}
\begin{minipage}{0.95\textwidth}
	\footnotesize
	\item[1] No CPU/GPU speed-up given for ResNet50.
	\item[2] CPU speed-ups for Batch-Shaping, ConvNet-AIG are from Batch-Shaping paper.
	\item[3] HRANK lacks implementation to compute CPU/GPU inference.
	\item[4] Only GPU speed-up reported for FPGM.
	\item[5] GPU speed-up for DynConv depends on a custom CUDA kernel.
	\item[*] GMAC values are simulated estimates.
\end{minipage}
		\end{tablenotes}
	\end{threeparttable}
\end{table}

\begin{table}[tb]
	\centering
	\setlength{\tabcolsep}{3pt}
	\renewcommand{\arraystretch}{1.3}
	\caption{Evaluation of standard and modified MobileNetV2 models on ImageNet. This table details top-1 accuracy, computational complexity (MMAC), model size (Params, in millions), and inference performance on CPU and GPU.}
	\label{imagenet_mobilenet_table}
	\begin{adjustbox}{max width=\textwidth}
		\begin{tabular}{lccccccc}
			\toprule
			\textbf{Method} & \textbf{Top-1} & \textbf{MMAC} & \textbf{Params} & \multicolumn{2}{c}{\textbf{Inference (ms)}} & \multicolumn{2}{c}{\textbf{Speed-up}} \\
			\cmidrule(lr){5-6} \cmidrule(lr){7-8}
			& \textbf{Acc (\%)} &  & \textbf{(M)} & \textbf{CPU} & \textbf{GPU} & \textbf{CPU} & \textbf{GPU} \\
			\midrule
			MobileNetV2 & 71.63 & 320.36 & 3.5 & 47.76 & 0.78 & {---} & {---} \\
			\rowcolor{gray!20} \textbf{MobileNetV2-MoD-L} & 71.55 & 344.76 & 3.34 & 42.94 & 0.71 & 1.11 & 1.10 \\
			MobileNetV2-MoD & 69.34 & 220.56 & 2.94 & 33.43 & 0.56 & 1.43 & 1.39 \\
			\bottomrule
		\end{tabular}
	\end{adjustbox}
\end{table}

\subsection{Semantic Segmentation on Cityscapes}

For semantic segmentation, we utilized the Cityscapes dataset \cite{2016_Cordts_CONF}, which contains high-quality, finely annotated images from 50 European cities, divided into 19 semantic classes (2,975 training, 500 validation, and 1,525 test images). The experiments were performed using three different random seeds. Implementation details are provided in Appendix \ref{implementation_details_cityscapes}.

Table \ref{tab:semantic_segmentation_cityscapes} presents the performance of Fully Convolutional Networks (FCNs) with ResNet-based MoD backbones for semantic segmentation on the Cityscapes validation dataset. The FCN with a ResNet86-MoD backbone matches the inference times of the standard FCN-R50 model while enhancing the mean Intersection over Union (mIoU) by 0.95\%, demonstrating MoD's ability to boost segmentation accuracy without additional computational costs. The FCN-R75-MoD provides similar accuracy as standard models but with computational efficiency improvements, achieving a 9\% speed-up on CPU and 7\% on GPU.

\begin{table}[tb]
	\centering
	\setlength{\tabcolsep}{3pt}
	\renewcommand{\arraystretch}{1.3}
\caption{Performance comparison of FCN models on the Cityscapes validation dataset, contrasting standard ResNet50 backbones against MoD versions. The table demonstrates that MoD models deliver performance comparable to or better than the standard FCN, with either reduced or similar inference times. Metrics include FLOPS (T), parameters (M), and inference durations (ms).}
	\label{tab:semantic_segmentation_cityscapes}
	\begin{adjustbox}{max width=\textwidth}
		\begin{tabular}{@{\extracolsep{\fill}}lccccccccc} 
			\toprule
			\textbf{Model} & \multicolumn{3}{c}{\textbf{Perf. Metrics (\%)}} & \textbf{FLOPS} & \textbf{Params} & \multicolumn{2}{c}{\textbf{Inference (ms)}} & \multicolumn{2}{c}{\textbf{Speed-up}} \\
			\cmidrule(lr){2-4} \cmidrule(lr){7-8} \cmidrule(lr){9-10}
			& \textbf{aAcc} & \textbf{mIoU} & \textbf{mAcc} & \textbf{(T)} & \textbf{(M)} & \textbf{CPU} & \textbf{GPU} & \textbf{CPU} & \textbf{GPU} \\
			\midrule
			\rowcolor{gray!20} \textbf{FCN-R86-MoD} & 95.75 & 73.75 & 81.75 & 0.377 & 47.15 & 8272.40 & 202.65 & 1.01 & 1.02 \\
			FCN-R50     & 95.68 & 72.80 & 80.25 & 0.393 & 47.11 & 8389.35 & 206.30 & {---} & {---} \\
			\rowcolor{gray!20} \textbf{FCN-R75-MoD} & 95.72 & 72.72 & 80.56 & 0.359 & 44.66 & 7687.20 & 192.50 & 1.09 & 1.07 \\
			FCN-R50-MoD & 95.37 & 71.58 & 79.54 & 0.322 & 39.66 & 6757.20 & 166.50 & 1.24 & 1.24 \\
			\bottomrule
		\end{tabular}
	\end{adjustbox}
\end{table}

\subsection{Object Detection on Pascal VOC}

For our object detection experiments, we utilized the PASCAL VOC dataset, employing the mmobjectdetection library \cite{2019_Chen} to configure and train our models. The experiments were performed using three different random seeds. Detailed implementation notes are provided in Appendix \ref{implementation_details_pascal_voc}.

The outcomes of our experiments with Faster R-CNN models on the PASCAL VOC dataset are presented in Table \ref{tab:object_detection_pascal_voc}. The Faster-RCNN with a ResNet86-MoD backbone demonstrates the MoD approach's effectiveness by achieving a 0.37\% improvement in mean Average Precision (mAP) and a 0.4\% increase in Average Precision at 50\% IoU threshold (AP50) compared to the standard Faster-RCNN-R50 model. Although GPU inference speed slightly decreases to 0.96x of the baseline, CPU processing speed is enhanced, showing a 1.10x improvement over the standard model. Similarly, the Faster-RCNN-R75-MoD model maintains accuracy close to the baseline while boosting inference efficiency, achieving an 11\% speed-up on CPU and 5\% on GPU.

\begin{table}[h]
	\centering
	\setlength{\tabcolsep}{3pt}
	\renewcommand{\arraystretch}{1.3}
\caption{Performance comparison of F-RCNN models on the Pascal VOC validation dataset, contrasting standard and ResNet-MoD backbones. The table shows that MoD variants match or exceed the standard F-RCNN's efficiency, achieving similar or better object detection performance with faster or equivalent inference times. Metrics include FLOPS (T), parameters (M), and inference durations (ms).}
	\label{tab:object_detection_pascal_voc}
	\begin{adjustbox}{max width=\textwidth}
		\begin{tabular}{@{\extracolsep{\fill}}lcccccccc} 
			\toprule
			\textbf{Model} & \multicolumn{2}{c}{\textbf{Perf. Metrics (\%)}} & \textbf{FLOPS} & \textbf{Params} & \multicolumn{2}{c}{\textbf{Inference (ms)}} & \multicolumn{2}{c}{\textbf{Speed-up}} \\
			\cmidrule(lr){2-3} \cmidrule(lr){6-7} \cmidrule(lr){8-9}
			& \textbf{mAP} & \textbf{AP50} & \textbf{(T)} & \textbf{(M)} & \textbf{CPU} & \textbf{GPU} & \textbf{CPU} & \textbf{GPU} \\
			\midrule
			\rowcolor{gray!20} \textbf{Faster-RCNN-R86-MoD} & 77.67 & 77.67 & 0.108 & 41.486 & 2813.9 & 83.5 & 1.10 & 0.96 \\
			Faster-RCNN-R50 & 77.30 & 77.27 & 0.111 & 41.446 & 3101.5 & 80.5 & {---} & {---} \\
			\rowcolor{gray!20} \textbf{Faster-RCNN-R75-MoD} & 77.04 & 77.03 & 0.104 & 38.989 & 2784.5 & 76.4 & 1.11 & 1.05 \\
			Faster-RCNN-R50-MoD & 75.10 & 75.10 & 0.949 & 33.994 & 2453.3 & 69.3 & 1.26 & 1.16 \\
			\bottomrule
		\end{tabular}
	\end{adjustbox}
\end{table}

\section{Channel Selection and Regularization Analysis}

\subsection{Channel Selector Mechanism}

\begin{figure}[tb]
	\centering
	\begin{subfigure}[b]{0.47\textwidth}
		\centering
		\includegraphics[width=\textwidth]{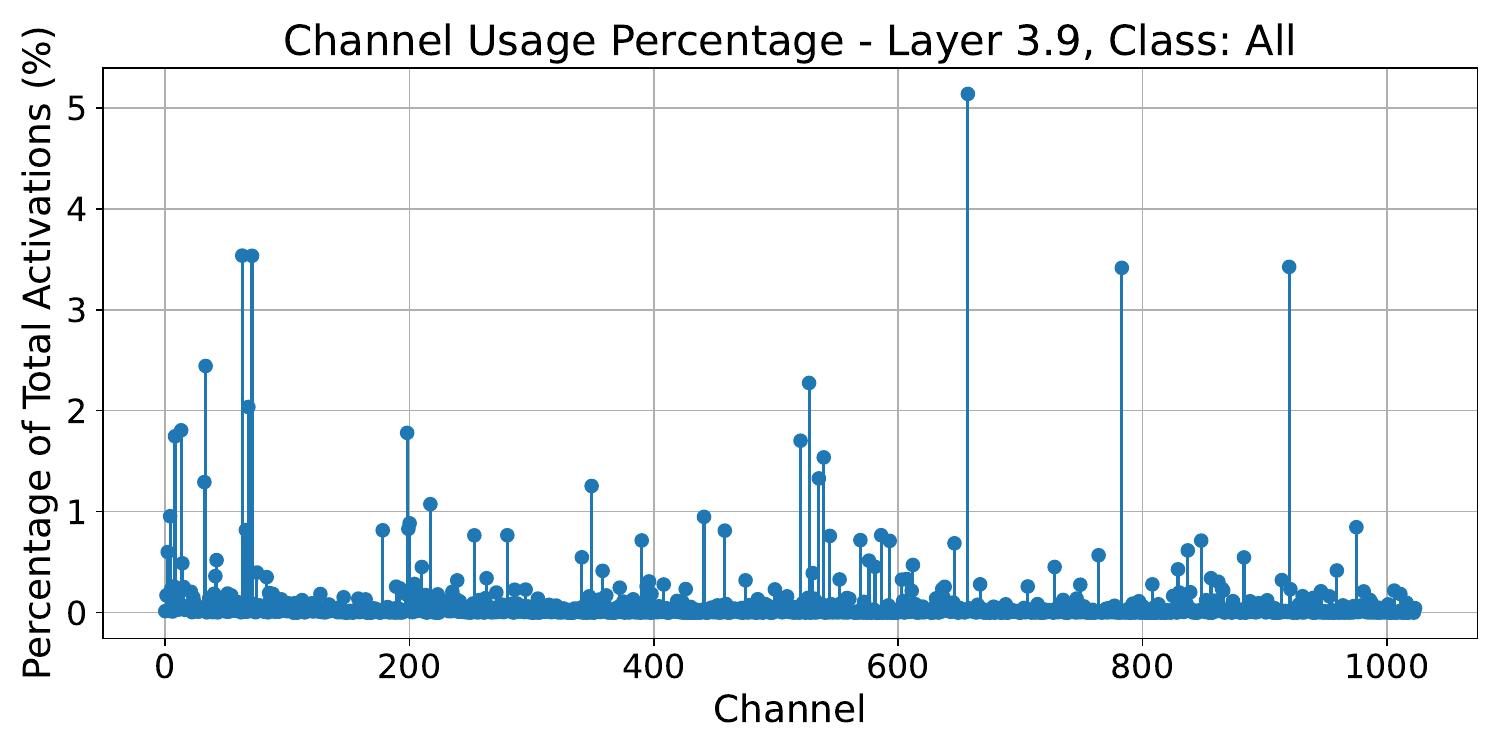}
		\caption{Class: All}
		\label{fig:class_all}
	\end{subfigure}
	\hfill
	\begin{subfigure}[b]{0.47\textwidth}
		\centering
		\includegraphics[width=\textwidth]{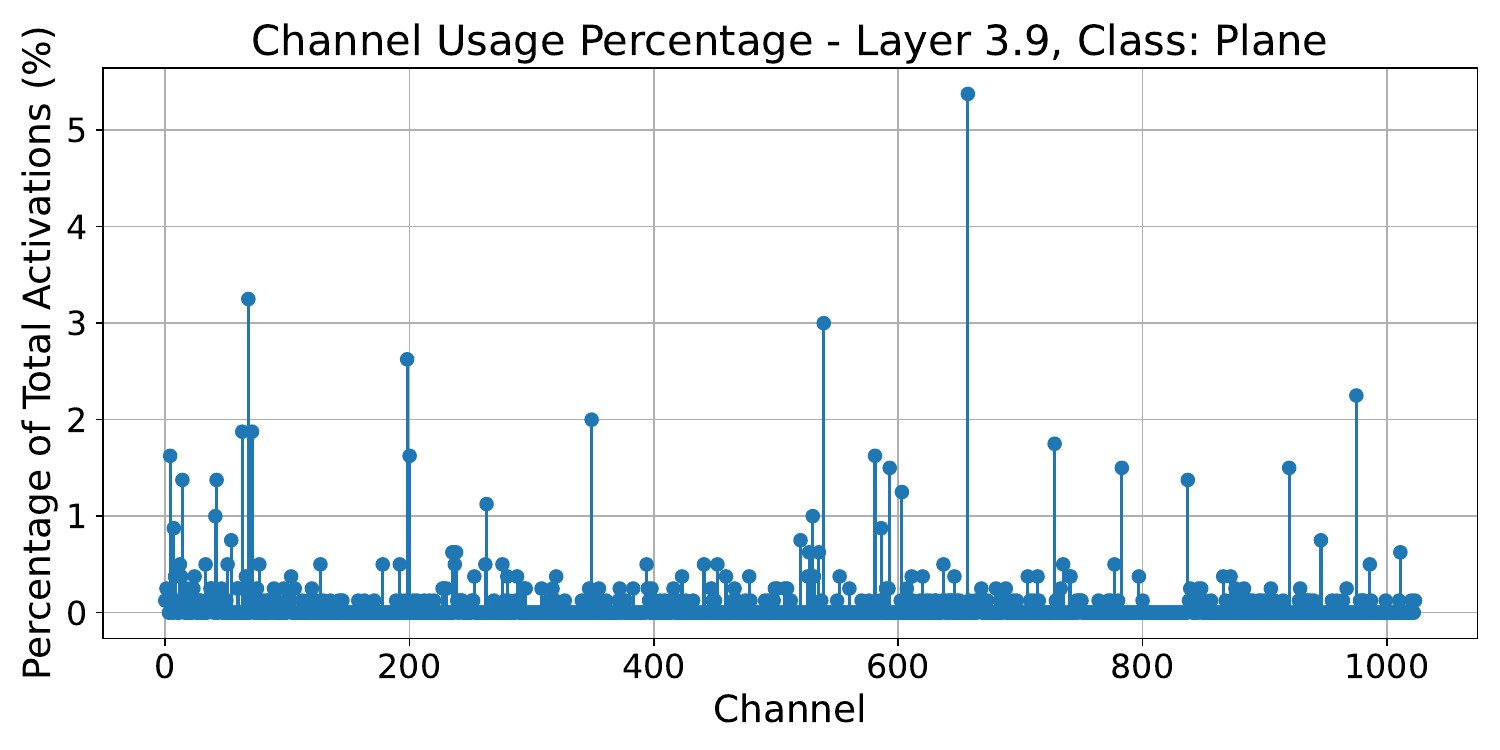}
		\caption{Class: Plane}
		\label{fig:class_plane}
	\end{subfigure}
	\vspace{1em}
	
	\begin{subfigure}[b]{0.47\textwidth}
		\centering
		\includegraphics[width=\textwidth]{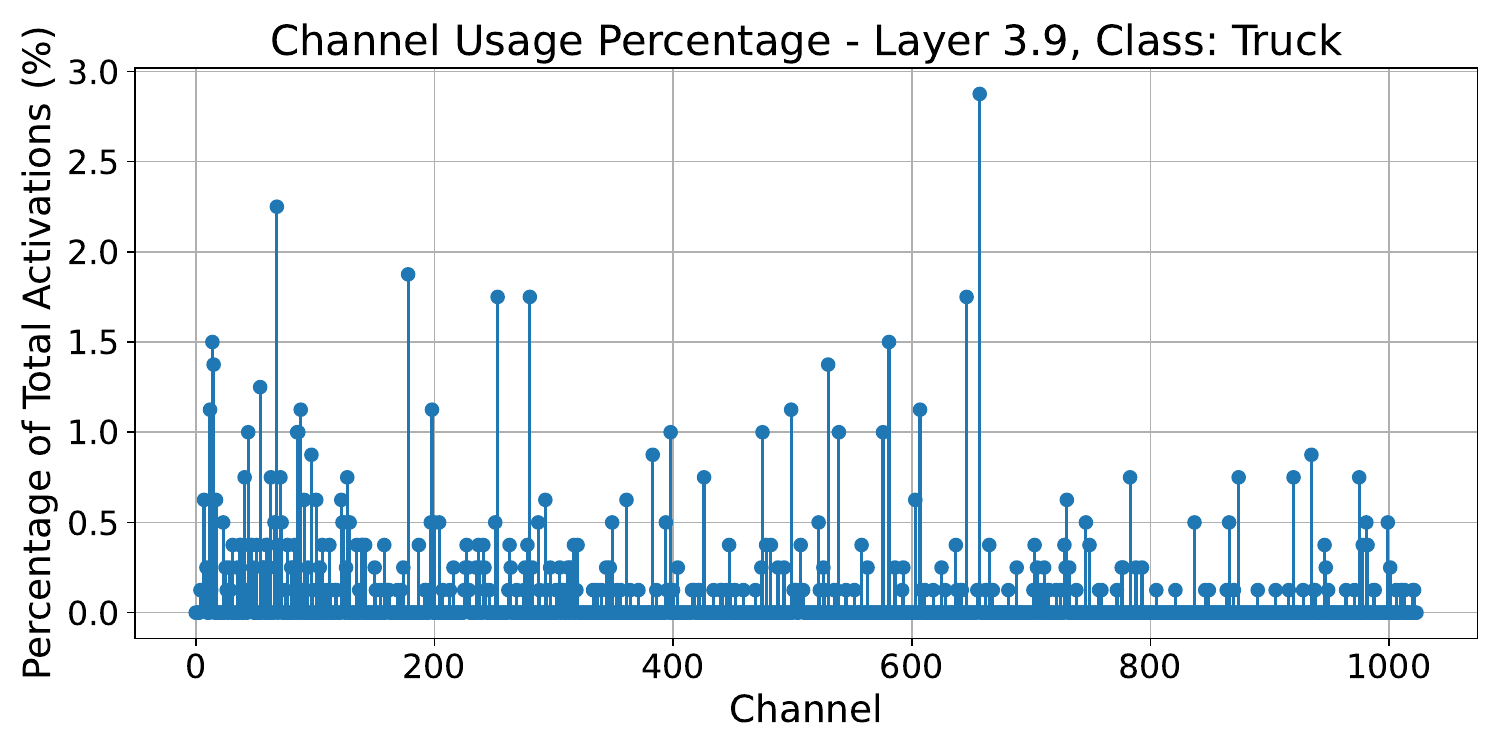}
		\caption{Class: Truck}
		\label{fig:class_truck}
	\end{subfigure}
	\hfill
	\begin{subfigure}[b]{0.47\textwidth}
		\centering
		\includegraphics[width=\textwidth]{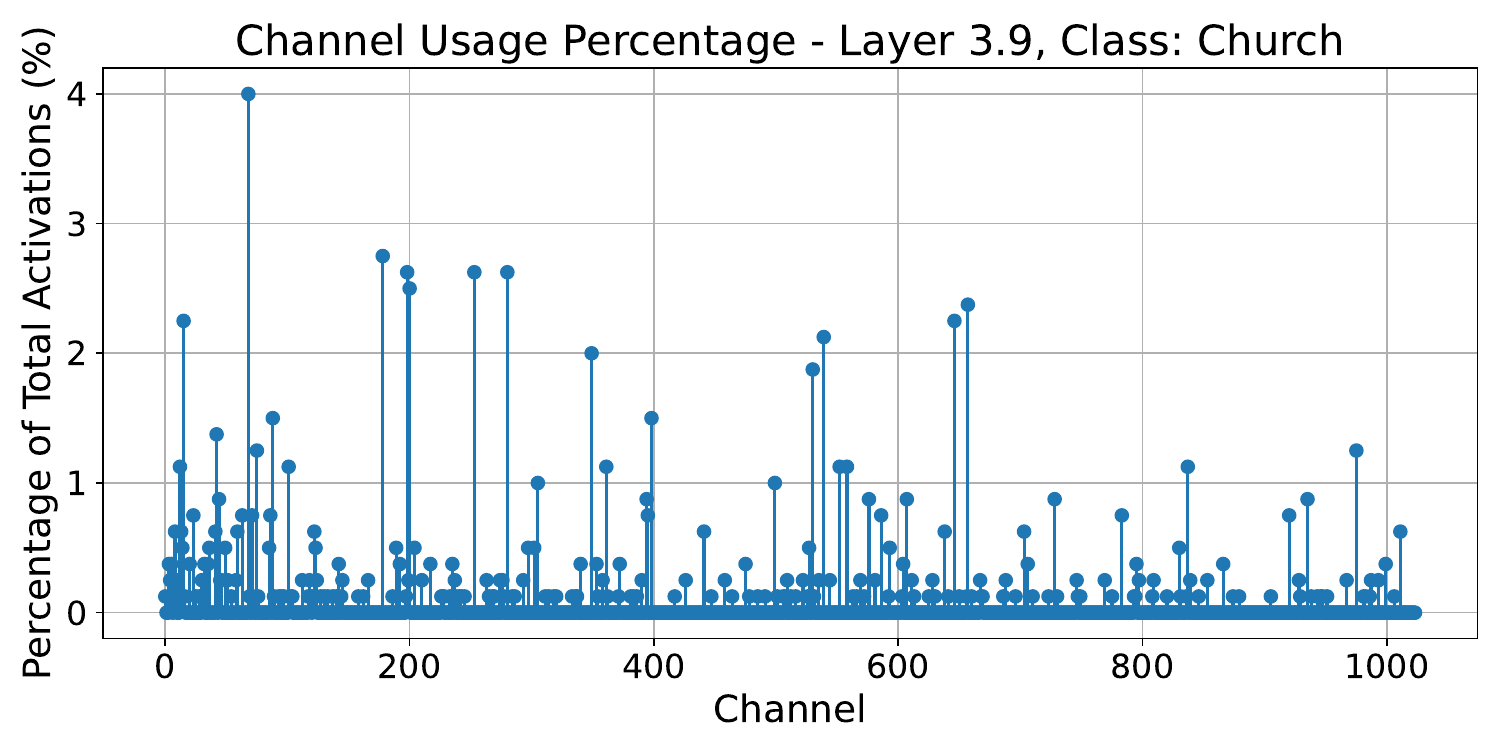}
		\caption{Class: Church}
		\label{fig:class_church}
	\end{subfigure}
	\vspace{1em}
	
	\begin{subfigure}[b]{0.47\textwidth}
		\centering
		\includegraphics[width=\textwidth]{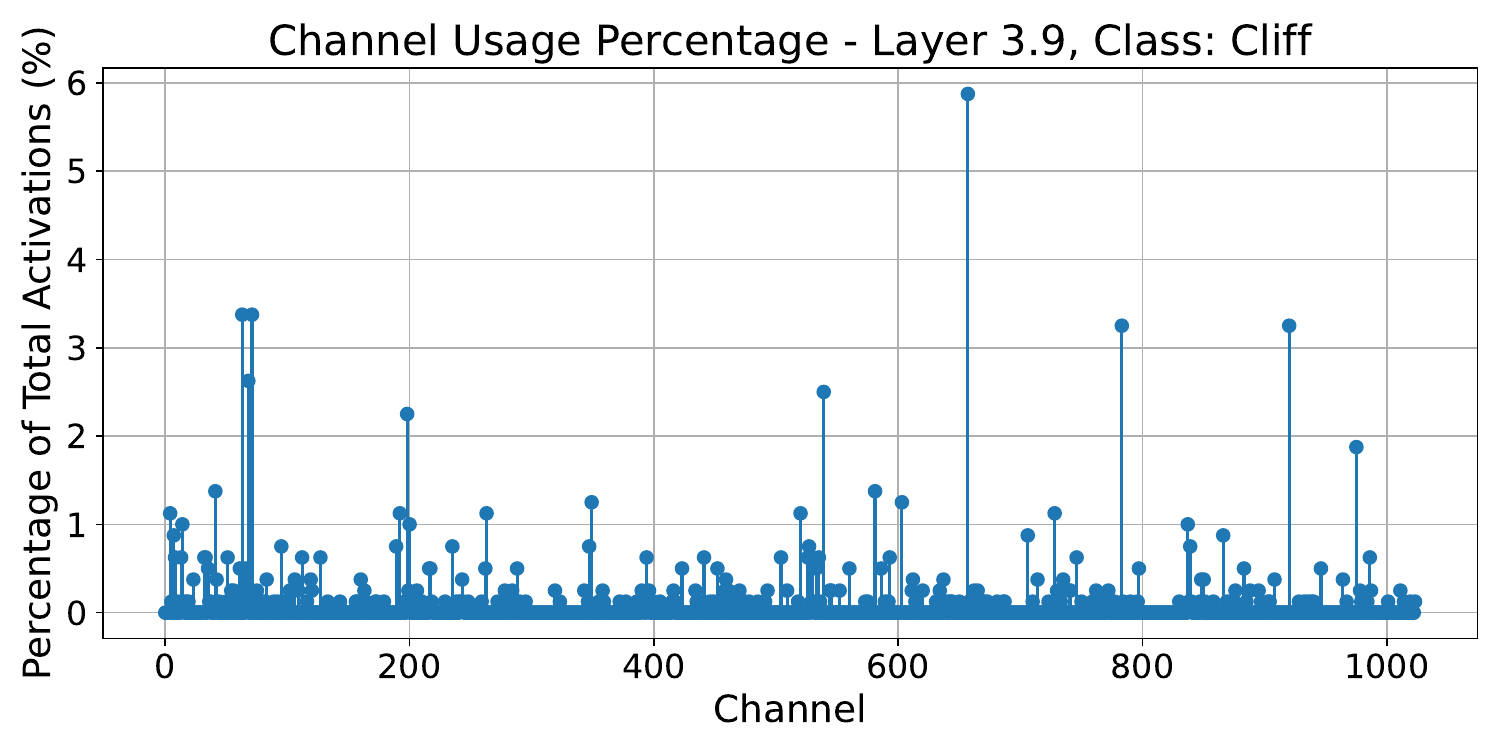}
		\caption{Class: Cliff}
		\label{fig:class_cliff}
	\end{subfigure}
	\hfill
	\begin{subfigure}[b]{0.47\textwidth}
		\centering
		\includegraphics[width=\textwidth]{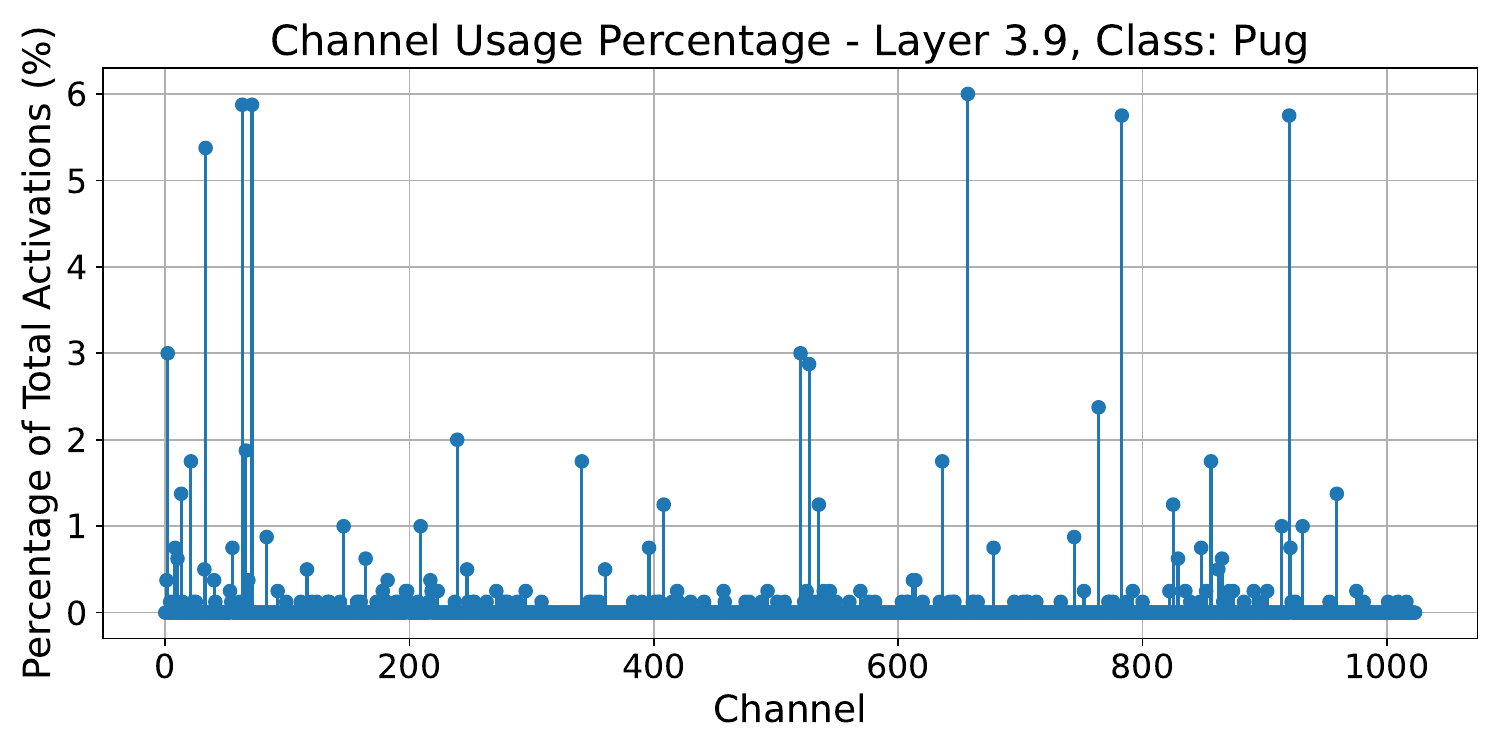}
		\caption{Class: Pug}
		\label{fig:class_pug}
	\end{subfigure}
	
\caption{Channel selection frequencies within the third module of a ResNet75-MoD for five diverse ImageNet classes: plane, truck, church, cliff, and pug. The analysis shows the percentage of times channels are selected by the Channel Selector out of the total selections in the layer, based on fifty samples per class. These percentages are compared to a baseline derived from all 1000 classes (\ref{fig:class_all}), indicating that the Channel Selector selects different channels for different classes.}
\label{fig:channel_selector}
\end{figure}

In this section, we investigate the practical operation of the Channel Selector within our CNN MoD approach. This analysis focuses on the Channel Selector within a MoD Block in the third module of a ResNet75-MoD. To provide a clearer understanding of the Channel Selector's behavior, we analyze how frequently channels are chosen in the top-$k$ routing mechanism. We use five different classes from the ImageNet validation set: plane, truck, church, cliff, and pug. For each class, fifty samples are drawn from the validation set, and we count the number of times each channel is selected. The percentage of occurrences for each channel is plotted in Figure \ref{fig:channel_selector}. For comparison, we also plot the percentages across all 1000 classes in Figure \ref{fig:class_all}.

Our analysis shows that the Channel Selector selects different channels for different classes, indicating that it adapts to enhance the detection of class-specific features. Additionally, some channels are selected more frequently across various classes, suggesting a potential for further computational load reduction through channel pruning techniques.
\subsection{Regularization Effect of Selective Processing}

An interesting byproduct of the CNN MoD approach is its regularization effect, caused by the selective processing of channels in the modified Conv-Blocks. Traditional CNN architectures typically process all channels within each Conv-Block, which can lead to learning overly specific features that may not generalize well to new data. In contrast, the MoD framework selectively reduces the number of channels processed, focusing on those most important ones for the current task.

This selective processing naturally encourages the network to prioritize and refine features that are more likely to be generalizable across various images. As a result, MoD acts as a form of natural regularization, pushing the network to extract more useful information from a limited set of inputs channels and thereby promoting the learning of robust, broadly applicable features. This is particularly advantageous for preventing overfitting, a common challenge in deep learning models, without additional explicit regularization strategies.

\section{Conclusion}

In this work, we presented CNN MoD, an approach inspired by the Mixture-of-Depths method developed for Transformers \cite{2024_Raposo}. This technique combines the advantages of static pruning and dynamic computing within a single framework. It optimizes computational resources by dynamically selecting key channels in feature maps for focused processing within the Conv-Blocks, while skipping less relevant channels.

It maintains a static computation graph, which optimizes the training and inference speed without the need for customized CUDA kernels, additional loss functions, or fine-tuning. These attributes separate MoD from other dynamic computing methods, offering a significant simplification of both training and inference processes. CNN MoD achieves performance comparable to traditional CNNs but with reduced inference times, GMACs, and parameters, or it surpasses them while maintaining similar inference times, GMACs, and parameters in image recognition, semantic segmentation, and object detection. For example, on ImageNet, ResNet86-MoD exceeds the performance of the standard ResNet50 by 0.45\% with a 6\% speed-up on CPU and 5\% on GPU. Moreover, ResNet75-MoD achieves the same performance as ResNet50 with a 25\% speed-up on CPU and 15\% on GPU.

While the fusion operation currently achieves significant efficiency gains, especially with high-dimensional datasets like ImageNet and real-world tasks such as Cityscapes and Pascal VOC, further optimization of the slicing operation could unlock even greater improvements. Future work will explore refining this component, potentially through a customized CUDA kernel. Further research into the optimal number of processed channels within layers is also promising for optimizing performance.

\FloatBarrier

\bibliographystyle{splncs04}
\bibliography{cnn_mod}

\appendix

\FloatBarrier

\newpage

\section{Implementation Details and Further Results on ImageNet} 
\label{appendix_imagenet}

\subsection{Implementation Details on ImageNet} \label{implementation_details_imagenet}

We performed our MoD experiments on ImageNet for 100 epochs, using PyTorch 1.10.1 \cite{2019_Paszke_CONF} across four Nvidia GeForce 1080Ti GPUs, following standard training protocols.\footnote{Refer to PyTorch's official training recipes at \href{https://github.com/pytorch/vision/tree/main/references/classification\#resnet}{PyTorch repository}.} Our configuration included a batch size of 256, an initial learning rate of 0.1 decreasing by a factor of 0.1 every 30 epochs, momentum set at 0.9, and weight decay of $1 \times 10^{-4}$. This setup aligns with the practices of comparable studies, such as DGNet \cite{2021_Li}.

\subsection{Architecture Details of ResNet MoD Models} \label{architecture_details_resnet_imagenet}

Table \ref{tab:resnet_mod_architectures} shows the layer configurations of the ResNet MoD models. The models incorporates alternating block patterns in each module.

\begin{table}[h]
	\centering
	\caption{Layer Configurations of ResNet MoD Models}
	\label{tab:resnet_mod_architectures}
	\begin{tabular}{|c|c|c|}
		\hline
		\textbf{Model} & \textbf{Layer Configuration} & \textbf{Block Type} \\
		\hline
		ResNet18-MoD & [2, 2, 2, 2] & Basic Block \\
		\hline
		ResNet26-MoD & [2, 2, 3, 4] & Basic Block \\
		\hline
		ResNet34-MoD & [3, 4, 6, 3] & Basic Block \\
		\hline
		ResNet42-MoD & [3, 3, 6, 6] & Basic Block \\
		\hline
		ResNet50-MoD & [3, 4, 6, 3] & Bottleneck Block \\
		\hline
		ResNet75-MoD & [3, 4, 14, 3] & Bottleneck Block \\
		\hline
		ResNet86-MoD & [3, 4, 18, 3] & Bottleneck Block \\
		\hline
		ResNet101-MoD & [3, 4, 23, 3] & Bottleneck Block \\
		\hline
		ResNet152-MoD & [3, 8, 36, 3] & Bottleneck Block \\
		\hline
	\end{tabular}%
\end{table}

\subsection{Architecture Details of MobileNetV2 MoD Models} 
\label{architecture_details_mobilenetv2_imagenet}

The MobileNetV2-MoD-L represents a deeper configuration of the standard MobileNetV2. The standard MobileNetV2 \cite{2018_Sandler_CONF} architecture utilizes a sequence of inverted residual blocks with a specific configuration pattern defined by the parameters \(t\) (expansion factor), \(c\) (number of channels), \(n\) (number of times the block is repeated), and \(s\) (stride). The MobileNetV2-MoD-L model modifies these parameters as shown in Table \ref{tab:mobile_net_mod_config}.

\begin{table}[h]
	\centering
	\caption{Comparison of Standard MobileNetV2 and MobileNetV2-MoD-L Configurations}
	\label{tab:mobile_net_mod_config}
	\begin{tabular}{|c|c|c|}
		\hline
		\textbf{Layer} & \textbf{Standard Configuration} & \textbf{MoD-L Configuration} \\
		\hline
		1 & [1, 16, 1, 1] & [1, 16, 1, 1] \\
					\hline
		2 & [6, 24, 2, 2] & [6, 32, 2, 2] \\
					\hline
		3 & [6, 32, 3, 2] & [6, 64, 3, 2] \\
					\hline
		4 & [6, 64, 4, 2] & [6, 96, 4, 2] \\
					\hline
		5 & [6, 96, 3, 1] & [6, 128, 3, 1] \\
					\hline
		6 & [6, 160, 3, 2] & [6, 160, 3, 2] \\
					\hline
		7 & [6, 320, 1, 1] & [6, 320, 1, 1] \\
		\hline
	\end{tabular}
\end{table}

\subsection{Comparative Performance of ResNet Models on ImageNet} \label{furter_results_imagenet}

Table \ref{imagenet_table_app} shows further results for MoD-enhanced ResNets, detailing improvements in computational efficiency, model compactness, and inference speeds.

\begin{table}[h]
	\centering
	\setlength{\tabcolsep}{3pt}
	\caption{Comparative performance of standard and ResNet-MoD models on the ImageNet dataset. The table evaluates top-1 accuracy, computational complexity (GMAC), model size (Params, in millions), and inference speed improvements on CPU and GPU.}
	\label{imagenet_table_app}
	
	\begin{adjustbox}{max width=\textwidth}
		\begin{tabular}{lccccccc}
			\toprule
			\textbf{Method} & \textbf{Top-1} & \textbf{GMAC} & \textbf{Params} & \multicolumn{2}{c}{\textbf{Inference (ms)}} & \multicolumn{2}{c}{\textbf{Speed-up}} \\
			\cmidrule(lr){5-6} \cmidrule(lr){7-8}
			& \textbf{Acc (\%)} &  & \textbf{(M)} & \textbf{CPU} & \textbf{GPU} & \textbf{CPU} & \textbf{GPU} \\
			\midrule
			ResNet34 & 73.92 & 3.68 & 21.29 & 98.83 & 1.27 & {---} & {---} \\
			ResNet42-MoD & 72.03 & 2.29 & 17.72 & 64.01 & 0.88 & 1.54 & 1.45 \\
			ResNet34-MoD & 71.44 & 2.06 & 12.93 & 60.79 & 0.83 & 1.63 & 1.53 \\
			\midrule
			ResNet18 & 70.37 & 1.82 & 11.18 & 53.86 & 0.75 & {---} & {---} \\
			ResNet26-MoD & 69.53 & 1.36 & 11.4 & 42.16 & 0.58 & 1.28 & 1.28 \\
			ResNet18-MoD & 64.05 & 0.89 & 5.46 & 33.63 & 0.47 & 1.60 & 1.58 \\
			\bottomrule
		\end{tabular}
	\end{adjustbox}

\end{table}

\subsection{Comparative Performance of ResNet MoD-l$k$ Models on ImageNet} \label{lastk_results_imagenet}

Table \ref{imagenet_table_app_lk} presents a comparison of the performance of standard and MoD-enhanced ResNet models with a specific focus on the position of reintegration of processed channels. MoD-l$k$ denotes models where processed channels are added to the last \(k\) channels of the original feature map. This experimental variation aims to assess the impact of consistent channel positioning on network performance. The results indicate that MoD-l$k$ models perform comparably to their counterparts where processed channels are added to the first \(k\) channels, suggesting that maintaining a consistent position for processed information within the feature maps is beneficial for optimizing model performance.

\begin{table}[h]
	\centering
	\setlength{\tabcolsep}{3pt}
\caption{This table presents a comparison between standard MoD models and the MoD-lk models, where processed channels are integrated into the last k channels. The evaluation covers top-1 accuracy, computational complexity (GMAC), model size (in millions of parameters), and inference speed improvements on both CPU and GPU. The aim is to analyze the impact of channel positioning on the performance of ResNet models on the ImageNet dataset.}
	\label{imagenet_table_app_lk}
	
	\begin{adjustbox}{max width=\textwidth}
		\begin{tabular}{lccccccc}
			\toprule
			\textbf{Method} & \textbf{Top-1} & \textbf{GMAC} & \textbf{Params} & \multicolumn{2}{c}{\textbf{Inference (ms)}} & \multicolumn{2}{c}{\textbf{Speed-up}} \\
			\cmidrule(lr){5-6} \cmidrule(lr){7-8}
			& \textbf{Acc (\%)} &  & \textbf{(M)} & \textbf{CPU} & \textbf{GPU} & \textbf{CPU} & \textbf{GPU} \\
			\midrule
			\rowcolor{gray!20} \textbf{ResNet86-MoD} & 76.72 & 3.92 & 25.60 & 150.96 & 2.40 & 1.06 & 1.05 \\
			ResNet86-MoD-lk & 76.64 & 3.92 & 25.60 & 150.96 & 2.40 & 1.06 & 1.05 \\
			\rowcolor{gray!20} \textbf{ResNet75-MoD} & 76.27 & 3.48 & 23.10 & 128.90 & 2.19 & 1.25 & 1.15 \\
			ResNet75-MoD-lk & 76.22 & 3.48 & 23.10 & 128.90 & 2.19 & 1.25 & 1.15 \\
			\rowcolor{gray!20} \textbf{ResNet50-MoD} & 74.79 & 2.60 & 18.11 & 108.74 & 1.75 & 1.48 & 1.44 \\
			ResNet50-MoD-lk & 74.57 & 2.60 & 18.11 & 108.74 & 1.75 & 1.48 & 1.44 \\
			\bottomrule
		\end{tabular}
	\end{adjustbox}
	
\end{table}

\subsection{Impact of Channel Parameter \(c\) and Integration Strategy on Accuracy and Inference Times}
\label{channel_param_effect}

In this section, we evaluate the influence of the channel parameter \(c\) on the top-1 validation accuracy and inference times of the ResNet50-MoD model on the ImageNet dataset. The parameter \(c\) determines the number of channels processed by the MoD approach. Additionally, we compare two strategies for reintegrating processed channels: adding them to the first \(k\) channels (S) versus adding them back to their original positions (OP).

As shown in Table \ref{tab:parameter_c}, reintegrating processed channels into the first \(k\) channels consistently outperforms the original position strategy, yielding better accuracy across all values of \(c\). Furthermore, using \( \mathbf{c = 64} \) has the optimal balance between accuracy and inference time. It should be noted that \(c > 64\) is not feasible since the number of channels in the first Conv-Block of ResNets is limited to 64.

\begin{table}[h!]
	\centering
	\setlength{\tabcolsep}{5pt}
	\caption{Top-1 accuracy and inference times in ms on ImageNet for different values of \(c\), comparing results when processed channels are added to the first \(k\) channels (S) versus their original positions (OP) for ResNet50-MoD.}
	\label{tab:parameter_c}
	\begin{tabular}{c c c c c c}
		\toprule
		\textbf{c} & \textbf{Top-1 (S)} & \textbf{Top-1 (OP)} & \textbf{GPU (S)} & \textbf{CPU (S)} \\
		\midrule
		2  & 72.51 & 55.36 & 2.23 & 132.64 \\
		4  & 73.74 & 60.78 & 1.99 & 118.85 \\
		8  & 74.43 & 63.05 & 1.90 & 114.85 \\
		16 & 74.74 & 65.35 & 1.97 & 114.06 \\
		32 & 74.68 & 70.80 & 1.83 & 111.30 \\
		64 & 74.79 & 70.31 & 1.75 & 108.74 \\
		\bottomrule
	\end{tabular}
\end{table}

\subsection{Impact of MoD on Performance Variance}
\label{performance_consistency}

We provide standard deviations for the ResNet experiments on ImageNet in Table \ref{tab:methods_comparison}. Other state-of-the-art pruning and dynamic computation methods, such as DGNet \cite{2021_Li}, Batch-Shaping \cite{2020_Bejnordi_CONF}, ConvNet-AIG \cite{2018_Veit_CONF}, HRANK \cite{2020_Lin_CONF}, FPGM \cite{2019_He_CONF}, and DynConv \cite{2020_Verelst_CONF}, do not report standard deviations, preventing direct comparison. Nevertheless, the consistent results across different MoD configurations demonstrate that the MoD approach does not introduce additional variance, as evidenced by the low standard deviations.

\begin{table}[h!]
	\centering
	\setlength{\tabcolsep}{5pt}
	\caption{Top-1 Accuracy and standard deviation comparison on ImageNet across various ResNet and ResNet-MoD models. This comparison shows that the MoD approach does not increase variance and maintains performance stability similar to standard ResNet models.}
	\label{tab:methods_comparison}
	\begin{tabular}{l c}
		\toprule
		\textbf{Method} & \textbf{Top-1 Acc. (\%) $\pm$ Std. Dev.} \\
		\midrule
		R152-MoD     & 77.81 $\pm$ 0.05 \\
		R101         & 77.81 $\pm$ 0.07 \\
		R101-MoD     & 77.08 $\pm$ 0.08 \\
		R86-MoD      & 76.72 $\pm$ 0.04 \\
		R75-MoD      & 76.27 $\pm$ 0.07 \\
		R50          & 76.25 $\pm$ 0.19 \\
		R50-MoD      & 74.79 $\pm$ 0.08 \\
		\bottomrule
	\end{tabular}
\end{table}

\section{Implementation Details Semantic Segmentation} \label{implementation_details_cityscapes}

In the Cityscapes experiments, PyTorch 1.10.1 \cite{2019_Paszke_CONF} and four Nvidia GeForce 1080Ti GPUs were used. Using the MMSegmentation Framework \cite{mmseg2020}, we utilized FCN \cite{2015_Long_CONF} on the dataset \cite{2016_Cordts_CONF}, which consists of 2,975 training, 500 validation, and 1,525 testing images across 19 semantic classes. Training involved resizing, random cropping, flipping, photometric distortion, normalization, and padding. Testing employed multi-scale flip augmentation and normalization.

The FCN model, with a ResNet50 backbone, used an Encoder-Decoder architecture with FCN-Head as the decode and auxiliary heads. The model used SyncBN and a dropout ratio of 0.1, with the auxiliary head contributing 40\% to the total loss.

Optimization was via SGD (learning rate 0.01, momentum 0.9, weight decay 0.0005). A polynomial decay learning rate policy was applied (power 0.9, minimum learning rate 1e-4), over 80,000 iterations with checkpoints and evaluations (focusing on mIoU) every 8,000 iterations.

\section{Implementation Details Object Detection} \label{implementation_details_pascal_voc}

\paragraph{Model Configuration:} We configure our Faster R-CNN \cite{2015_Ren_CONF} with a ResNet-50 backbone and a Feature Pyramid Network (FPN) neck for multi-scale feature extraction.

\paragraph{Data Preprocessing and Augmentation:} Our preprocessing pipeline employs a sequence of transformations to prepare input images for object detection tasks. Initially, images are loaded and their corresponding annotations are retrieved. Subsequently, we resize the images to a resolution of $(1000, 600)$, ensuring the preservation of their original aspect ratio. To augment the dataset and introduce variability, we apply random horizontal flips with a 50\% probability. This augmentation strategy is applied uniformly across the training dataset, aiming to enhance model robustness and generalization capability. For validation, images undergo a similar resizing process without the application of random flips, maintaining consistency in evaluation conditions. 

\paragraph{Training Configuration:} The model is trained on the combined train sets of VOC2007 and VOC2012 and evaluated on the VOC2007 val set. Training uses a batch size of 2. We adopt SGD with momentum and weight decay, adjusting the learning rate as per a predefined schedule. The mean Average Precision (mAP) metric, calculated using the ``11points'' interpolation method, serves as the evaluation metric.

\paragraph{Evaluation:} The evaluation on the VOC2007 val set employs the standard VOC mAP metric, adhering to the ``11points'' method. This setup mirrors the training configuration but without data augmentation, ensuring deterministic inference.

\section{Experiments on CIFAR} \label{exp_CIFAR}

Our evaluation of the MoD approach was performed on the CIFAR-10/100 datasets, comprising 50,000 training and 10,000 test color images of 32x32 pixels. We utilized a range of CNN architectures for our experiments, including ResNet18/34/50 \cite{2016_He_CONF} and VGG16/19-BN \cite{2015_Simonyan_CONF}. Table \ref{table_cifar} presents the results of applying the MoD approach to the CNNs trained on CIFAR-10/100.

To ensure robustness and reproducibility, each model was trained and evaluated five times using different random seeds, impacting network initialization, data ordering, and augmentation processes. We present the mean test accuracy and its standard deviation for these trials. The data split comprised 90\% for training and 10\% for validation, with the best-performing model on the validation set chosen for the final evaluation.

\begin{table}[h]
	\centering
	\setlength{\tabcolsep}{3pt}
	\renewcommand{\arraystretch}{1.3}
	\caption{Performance metrics comparison on the CIFAR-10 and CIFAR-100 datasets using standard models and their MoD variants. This table illustrates the trade-off between efficiency and inference time, demonstrating that the MoD models can achieve comparable performance to the standard models at faster inference times or improved performance at comparable inference times. FLOPS are in millions of multiply-accumulate operations (MMAC), parameters in millions (M), and inference times in milliseconds (ms).} \label{table_cifar}
	\begin{adjustbox}{max width=\textwidth}
		\begin{tabular}{@{\extracolsep{\fill}}lcccccccc}
			\toprule
			\textbf{Model} & \textbf{Set} & \textbf{Test Acc.} & \textbf{FLOPS} & \textbf{Params} & \multicolumn{2}{c}{\textbf{Inference (ms)}} & \multicolumn{2}{c}{\textbf{Speed-up}} \\
			\cmidrule(lr){6-7} \cmidrule(lr){8-9}
			& & \textbf{(\%)} & \textbf{(MMAC)} & \textbf{(M)} & \textbf{CPU} & \textbf{GPU} & \textbf{CPU} & \textbf{GPU} \\
			\midrule
			ResNet18 & C10 & $94.04 \pm 0.08$ & 557 & 11.17 & 15.67 & 0.24 & - & - \\
			ResNet18-MoD & C10 & $92.37 \pm 0.18$ & 255 & 4.95 & 8.32 & 0.14 & 1.88 & 1.73 \\
			ResNet34 & C10 & $93.69 \pm 0.27$ & 1016 & 21.28 & 30.52 & 0.42 & - & - \\
			ResNet34-MoD & C10 & $93.83 \pm 0.20$ & 633 & 12.42 & 18.10 & 0.27 & 1.69 & 1.54 \\
			ResNet50 & C10 & $93.31 \pm 0.33$ & 1310 & 23.52 & 48.47 & 0.83 & - & - \\
			ResNet50-MoD & C10 & $93.24 \pm 0.24$ & 808 & 16.07 & 33.76 & 0.58 & 1.44 & 1.41 \\
			\midrule
			ResNet18 & C100 & $76.47 \pm 0.18$ & 557 & 11.22 & 14.99 & 0.23 & - & - \\
			ResNet18-MoD & C100 & $72.73 \pm 0.21$ & 255 & 4.99 & 8.89 & 0.13 & 1.69 & 1.75 \\
			ResNet34 & C100 & $77.07 \pm 0.41$ & 1160 & 21.33 & 30.67 & 0.42 & - & - \\
			ResNet34-MoD & C100 & $76.86 \pm 0.23$ & 633 & 12.47 & 18.66 & 0.27 & 1.64 & 1.55 \\
			ResNet50 & C100 & $76.17 \pm 0.63$ & 1310 & 23.71 & 46.64 & 0.81 & - & - \\
			ResNet50-MoD & C100 & $76.76 \pm 0.61$ & 808 & 16.26 & 34.81 & 0.58 & 1.34 & 1.41 \\
			\midrule
			VGG16-BN & C10 & $93.27 \pm 0.11$ & 315 & 15.25 & 8.80 & 0.14 & - & - \\
			VGG16-BN-MoD & C10 & $91.79 \pm 0.14$ & 155 & 9.83 & 5.47 & 0.16 & 1.61 & 0.87 \\
			VGG19-BN & C10 & $93.21 \pm 0.07$ & 400 & 20.57 & 11.55 & 0.18 & - & - \\
			VGG19-BN-MoD & C10 & $91.82 \pm 0.18$ & 155 & 9.91 & 5.81 & 0.22 & 1.99 & 0.81 \\
			\midrule
			VGG16-BN & C100 & $72.48 \pm 0.32$ & 315 & 15.30 & 9.27 & 0.14 & - & - \\
			VGG16-BN-MoD & C100 & $69.23 \pm 0.25$ & 155 & 9.88 & 6.41 & 0.15 & 1.45 & 0.95 \\
			VGG19-BN & C100 & $71.34 \pm 0.12$ & 400 & 20.61 & 11.38 & 0.18 & - & - \\
			VGG19-BN-MoD & C100 & $69.19 \pm 0.11$ & 155 & 9.96 & 5.75 & 0.23 & 1.98 & 0.80 \\
			\bottomrule
		\end{tabular}
	\end{adjustbox}
\end{table}

\section{Comparative Analysis of MoD in CNNs and Transformers}
\label{sec:comparison_mod}

In the main body of this paper, we detailed the application of the MoD approach to CNNs. This section aims to outline how this approach differs from the Mixture-of-Depths application in Transformers \cite{2024_Raposo}. \\
\\
\textbf{Token vs. Channel Processing:}
\begin{itemize}
	\item \textbf{Transformers:} In Transformers, MoD operates at the token level. Tokens represent the units of data processed throughout the model's architecture, typically as subwords or whole words. They are processed throughout the entire Transformer architecture.
	
	\item \textbf{CNNs:} Conversely, MoD in CNNs treats channels within feature maps as ``tokens''. This novel approach differs from Transformers because channels in traditional CNNs are not treated as tokens, and their significance and composition vary from one convolutional layer to the next.
\end{itemize}
\textbf{Token vs. Channel Selection:}
\begin{itemize}
	\item \textbf{Transformers:} Selection is based on a linear projection that assigns a scalar value to each token.
	\item \textbf{CNNs:} CNNs use a mini neural network (incorporating Adaptive Average Pooling 2D, a two-layer fully connected network with Sigmoid activation) inspired by Squeeze-and-Excitation blocks \cite{2018_Hu_CONF}, specifically tailored for image-based tasks.
\end{itemize}
\textbf{Architecture Modification:}
\begin{itemize}
	\item \textbf{Transformers:} The architecture of Transformer blocks remains unchanged with the use of MoD; only the quantity of processed tokens varies. Transformer models are designed to handle a variable number of tokens.
	\item \textbf{CNNs:} In CNNs, varying the number of channels in convolutional layers is impractical. MoD thus requires adjustments to the convolutional layers themselves, including a reduction in the number of channels in the convolution kernels to match the reduced number of input feature map channels.
\end{itemize}
\textbf{Reintegration of Processed Information:}
\begin{itemize}
	\item \textbf{Transformers:} Processed tokens are added back to their original counterparts, a method made effective through the use of positional encoding.
	\item \textbf{CNNs:} Unlike in Transformers, neither replacing nor adding back processed channels to their original positions has proven effective in CNNs. More effective is the addition of processed channels to a fixed set of channels, such as the first \(k\) channels, to maintain consistency in locating processed information within the network.
\end{itemize}

\end{document}